\documentclass[letterpaper]{article} 
\usepackage{aaai25}  
\usepackage{times}  
\usepackage{helvet}  
\usepackage{xspace}

\usepackage{courier}  
\usepackage[hyphens]{url}  
\usepackage{graphicx} 
\usepackage{natbib}  
\usepackage{multirow}
\usepackage{float}
\usepackage{caption} 
\usepackage{algorithm}
\usepackage{algorithmic}
\usepackage{amsmath}
\usepackage{amssymb}
\usepackage{soul}
\usepackage{cancel}

\usepackage{newfloat}
\usepackage{xcolor}
\usepackage{listings}
\DeclareCaptionStyle{ruled}{labelfont=normalfont,labelsep=colon,strut=off} 
\floatstyle{ruled}
\newfloat{listing}{tb}{lst}{}
\floatname{listing}{Listing}
\newcommand{\ie}{\emph{i.e.}\xspace} 
\newcommand{\eg}{\emph{e.g.}\xspace} 

\newcommand{\our}{\textsc{Flight2Vec}\xspace}

\title{Effective and Efficient Representation Learning for Flight Trajectories}
\author{
    Shuo Liu\textsuperscript{\rm 1,2}, Wenbin Li\textsuperscript{\rm 2,3}, Di Yao\textsuperscript{\rm 2\raisebox{-0.5ex}{*}}, Jingping Bi\textsuperscript{\rm 2}\thanks{Corresponding authors.}
}
\affiliations{
    \textsuperscript{\rm 1}School of Advanced Interdisciplinary Sciences, University of Chinese Academy of Sciences, China\\
    \textsuperscript{\rm 2}Institute of Computing Technology, Chinese Academy of Sciences, China\\
    \textsuperscript{\rm 3}University of Chinese Academy of Sciences, China\\

    \{liushuo22s,liwenbin20z,yaodi,bjp\}@ict.ac.cn\\
}

\begin{document}
\maketitle
\begin{abstract}
Flight trajectory data plays a vital role in the traffic management community, especially for downstream tasks such as trajectory prediction, flight recognition, and anomaly detection. Existing works often utilize handcrafted features and design models for different tasks individually, which heavily rely on domain expertise and are hard to extend. We argue that different flight analysis tasks share the same useful features of the trajectory. Jointly learning a unified representation for flight trajectories could be beneficial for improving the performance of various tasks. However, flight trajectory representation learning (TRL) faces two primary challenges, \ie unbalanced behavior density and 3D spatial continuity, which disable recent general TRL methods. 
In this paper, we propose \our, a flight-specific representation learning method to address these challenges. Specifically, a behavior-adaptive patching mechanism is used to inspire the learned representation to pay more attention to behavior-dense segments. Moreover, we introduce a motion trend learning technique that guides the model to memorize not only the precise locations, but also the motion trend to generate better representations.  
Extensive experimental results demonstrate that \our significantly improves performance in downstream tasks such as flight trajectory prediction, flight recognition, and anomaly detection.
\end{abstract}
\section{Introduction}
With the continual development of air transportation, flights tend to broadcast their current locations for safe and efficient air traffic control~\cite{zhang2023flight}. The collected flight trajectories are one of the most critical data sources for air traffic management, attracting increasing attention from both academic and industrial fields. Recent works show that deep representation learning has become the dominant technique and achieved significant performance for various flight trajectory-related tasks, such as trajectory prediction~\cite{liu2018predicting,wu2022long,zhang2023flight,guo2024flightbert++}, flight monitoring~\cite{zhang2016data,fernandez2019flight}, and anomaly detection~\cite{olive2020detection,memarzadeh2022multiclass,memarzadeh2023anomaly}. However, these works either utilize handcrafted features or design representation models for one specified task. We argue that different flight tasks may share the same useful features of trajectory. Jointly learning a unified representation for flight trajectories could be beneficial for boosting the performance of various tasks, which motivates this work. 

\begin{figure}[t]
\centering
\includegraphics[width=1\columnwidth]{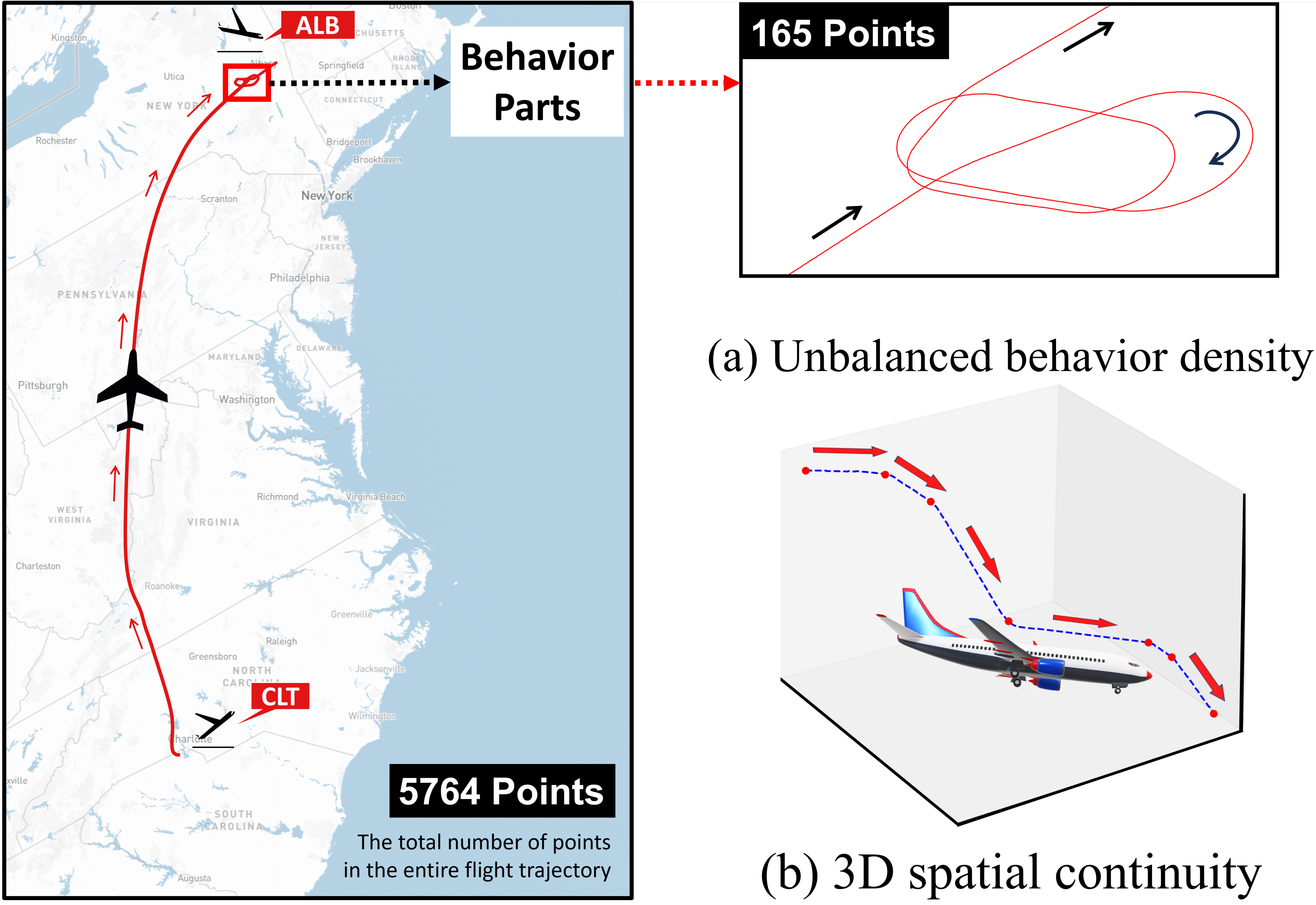} 
\vspace{-3ex}
\caption{The motivation of \our}
\label{motivation}
\vspace{-4ex}
\end{figure}

Learning unified trajectory representation for different tasks has been well-studied on vehicle and human mobility trajectories~\cite{yao2017trajectory,li2018deep,chen2021robust,jiang2023self}. Deep sequential models, such as Recurrent Neural Networks~\cite{yao2017trajectory,li2018deep} and Transformer~\cite{chen2021robust,yao2022trajgat,jiang2023self} are employed to encode the spatial-temporal correlations and transform raw trajectories into generic representation vectors. Nevertheless, existing solutions are hard to extend to represent flight trajectories due to the two challenges, \ie unbalanced behavior density and 3D spatial continuity.

\textbf{Unbalanced behavior density.} Flight trajectories are characterized by high point density but sparse behavior information. As shown in Figure~\ref{motivation}(a), the aircraft usually fly in a straight line with a fixed attitude. The representative activities, such as turns, holding patterns, and takeoff/landing phases, only account for a relatively small part, \eg $5\%$,  of the whole trajectory. Existing TRL methods usually treat every point equally without considering the density of behaviors, leading to inconsequential representations.

\textbf{3D spatial continuity.} Flights move in a three-dimensional space, exhibiting more complex spatial patterns compared to ground trajectories. As shown in Figure~\ref{motivation}(b), the movement of fights follows the spatial proximity in 3D space, \ie the motion trend of flights are constrained in the forward direction. However, previous works utilize MSE loss to reconstruct location coordinates, which does not effectively capture the spatial continuity and dependencies. Moreover, MSE loss is sensitive to outliers, which significantly affect the representation learning process. 

In this paper, we propose a generic flight trajectory representation learning method \our which encourages the learned representation paying more attention to behavior parts and memorize the motion trend of flights. Specifically, a novel behavior-based patching mechanism is designed to automatically sample trajectories according to the behavior density and reduce the sequence length of model input. \our first employs a criterion to select the behavior parts of trajectories and construct patches for each behavior respectively. The generated patches are feed to the decoder-only Transformer to obtain the representations. To model the 3D spatial continuity, we propose a motion trend learning technique which predicts the moving direction of each record as a $26$-classes classification task. Along with the MSE loss, \our can not only extract the information of spatial coordinates but also capture the motion trend of flights. In summary, the main contributions of this paper are summarized as follows:
\begin{itemize}
    \item We propose an effective and efficient representation learning framework \our for flight trajectories. To the best of our knowledge, this is the first work that designs a general representation learning framework specifically tailored to the characteristics of flight trajectories.
    
    \item We introduce an behavior-adaptive patching mechanism to achieve effective trajectory representation learning while preserving the information of behavior density. Additionally, we propose a motion trend learning technique that explicitly models the spatial continuity flight trajectories without requiring additional features.
    
    \item Extensive experiments demonstrate that \our significantly enhances performance across various downstream tasks, such as anomaly detection, trajectory prediction, and flight monitoring.
\end{itemize}

\section{Related Work}
\subsubsection{Flight Trajectory Analysis Framework.}
Existing research on flight trajectory data analysis always requires manual feature engineering and specialized models for each task, heavily relying on domain expertise and being difficult to extend. For instance,~\cite{guo2022flightbert,guo2024flightbert++} proposed a feature representation method based on binary encoding (BE) for trajectory prediction, utilizing Conv1D and Transformer modules to capture spatiotemporal features of trajectory points.~\cite{fang2021symbolic} selected 10 correlation coefficient features, such as Oil Temperature (OilT) and Oil Pressure, for trajectory recognition.~\cite{qin2022flight} introduced unsupervised feature engineering methods to map input data into latent feature spaces for anomaly detection. The limitation of these methods is that they require expert knowledge and complex feature engineering to extract useful information. There are also some learning-based methods, but they do not solve the problem of uneven behavior density. They either model the entire trajectory indiscriminately~\cite{guo2024flightbert++} or only focus on specific segments such as takeoff and landing phases~\cite{fernandez2019flight,memarzadeh2022multiclass}, unable to adaptively identify and model informative segments of the trajectory. In conclusion, although existing methods have achieved success in specific tasks, they rely on complex feature engineering highlights the need for a unified approach.

\subsubsection{Trajectory Representation Learning.}
Trajectory Representation Learning (TRL) has gained significant attention in the data engineering community due to its effectiveness in enhancing various downstream tasks. Existing methods can be broadly divided into two categories: road network-based methods and grid-based methods. Road network-based methods~\cite{fu2020trembr,chen2021robust,jiang2023self,qian2024context} map trajectories to nodes of the road network and learn representations of nodes to generate trajectory representations. However, flight trajectories can move in three-dimensional space without the constraints of road networks, making road network-based methods inapplicable. Grid-based methods~\cite{yao2018learning,yao2019computing,li2018deep,yang2021t3s,jing2022can} divide the geographical space into a grid of cells and map trajectories to these grids to learn representations. These methods are also difficult to apply to flight trajectories since dividing 3D space into grids will result in an exponential increase in the number of grids. This will make the trajectory data on many grids too sparse, thus the model cannot effectively learn trajectory representations. Overall, there is no existing work specifically focused on representation learning for flight trajectories and existing TRL methods are not applicable to flight trajectories.

\subsubsection{Patch-based Transformer.}
In recent years, patch-based approaches for time series analysis have emerged as mainstream. 
Compared to point-wise processing methods, these models not only improve efficiency but also exhibit notable performance improvements, revealing the importance of enhancing modeling of local semantics through patching.
For instance, PatchTST~\cite{nie2022time} segments each time series into patches and employs an Multi-layer Perceptron (MLP) to feed patch embeddings into a Transformer. This method also utilizes mask-based unsupervised pre-training to learn representations that can generalize to various downstream tasks. Following PatchTST, a series of patch-based Transformer models for time series have continually set new benchmarks in various downstream tasks, such as prediction~\cite{wang2024timemixer,chen2024multi}, classification~\cite{li2024time} and anomaly detection~\cite{yang2023dcdetector}. 
Notably, HDMixer~\cite{huang2024hdmixer} has revealing the importance of patch division. This research shows that incorrect patch boundaries can obscure local patterns and disrupt the semantic continuity of the sequence. 
However, these methods do not address the issue of unbalanced behavior density, \ie, they fail to effectively capture sparse but critical behaviors in flight trajectories. 

\section{Preliminary}
In this section, we first define the problem and then describe the proposed method \our, respectively. 

\begin{figure}[t]
\centering
\includegraphics[width=1\columnwidth]{figure/framework.pdf} 
\caption{Overview of \our}
\vspace{-4ex}
\label{fig:overview}
\end{figure}

\subsubsection{Problem Definition.}
Given a flight trajectory denoted by \( T = \{x_1, x_2, \ldots, x_n\} \), where each $x_i$ represents a recorded point at timestamp $i$, the objective of \textit{Trajectory Representation Learning} (TRL) is to generate a general low-dimensional representation, \( \mathbf{v} \in \mathbb{R}^D \), that can be utilized in various downstream tasks.

In this study, each trajectory point $x_i$ is described by six key attributes related to the aircraft's status:

\begin{equation*}
x_t = [lon_i, lat_i, alt_i, V_{lon_i}, V_{lat_i}, V_{alt_i}],
\end{equation*}
where \( lon_i \), \( lat_i \), and \( alt_i \) represent the longitude, latitude, and altitude of the aircraft, respectively. The attributes \( V_{lon_i} \), \( V_{lat_i} \), and \( V_{alt_i} \) denote the velocity components in the longitudinal, latitudinal, and altitude dimensions, respectively.

\subsubsection{Overview of \our.}
As shown in Figure~\ref{fig:overview}, \our consists of two key components, \ie behavior-adaptive patching Transformer and model optimization. In the first component, we segment the flight trajectory into a sequence of patches according to the density of behaviors. For non-behavior segments, we down-sample the orginal records and compress the long segments into patches with equal size. The generated patches are feed to a patch Transformer encoder to obtain the flight representation. To optimize the parameters in \our, a motion trend learning approach is proposed along with the traditional MSE loss to reconstruct the masked patches. It encourages the learned representations memorize not only the location coordinates but also the moving directions of each records in the masked patches.

\section{Methodology}
We specify the two key components of \our, \ie, the behavior-adaptive patching Transformer, and the optimization of the model, respectively.

\subsection{Behavior-Adaptive Patching Transformer}

Behaviors such as takeoff and turning only account for a very small part of flight trajectories, but they are informative and crucial for modeling trajectory patterns.
Therefore, we propose an behavior-based patching mechanism that extracts more informative features by amplifying behavior-dense segments.

\subsubsection{Behavior-Based Patching}
\begin{figure}[t]
\centering
\includegraphics[width=0.8\columnwidth]{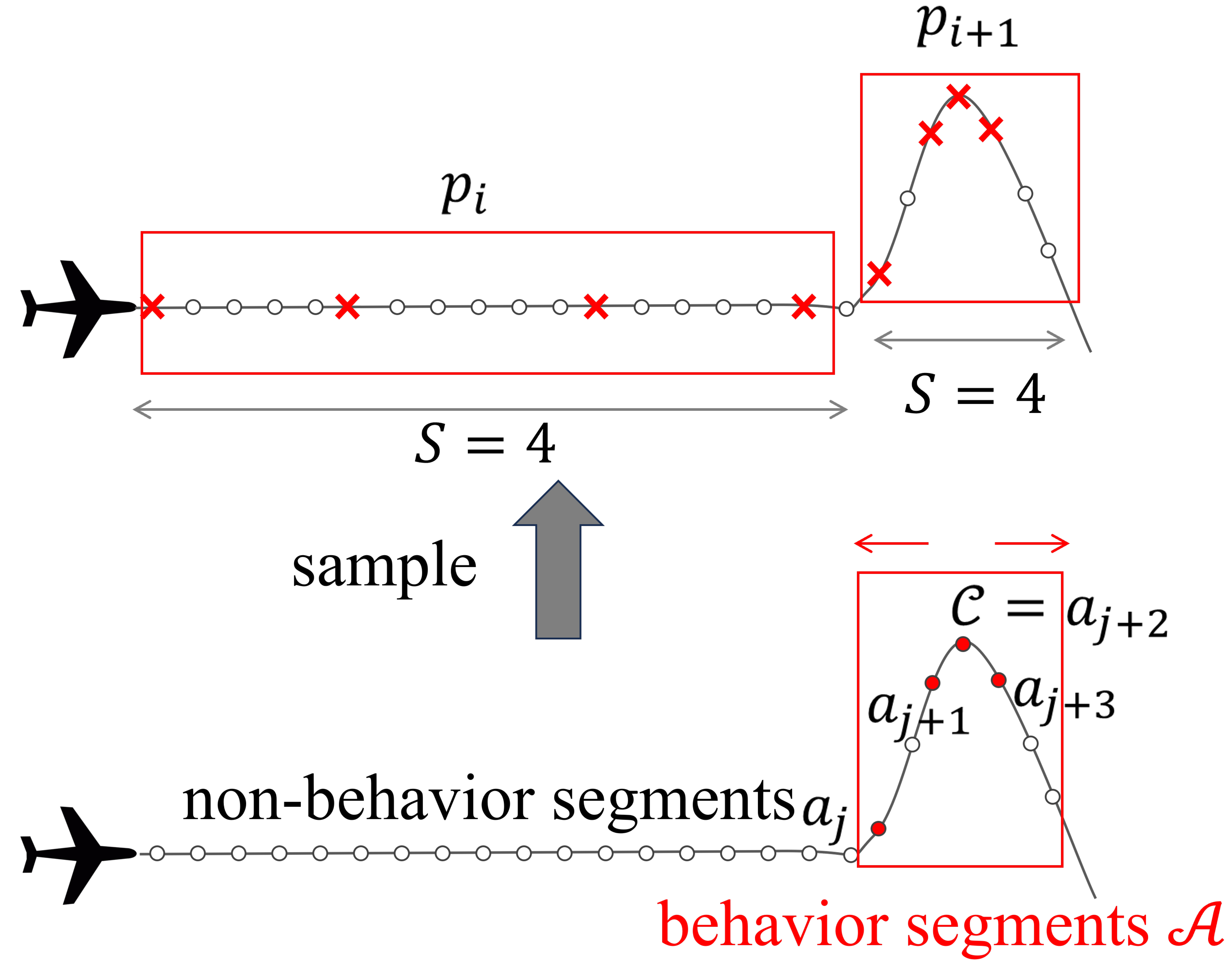} 
\caption{Behavior-Based Patching}
\label{fig:pacthing}
\vspace{-3ex}
\end{figure}

We observe that: (i) behavior segments reflect significant trajectory changes, such as turns and holding patterns; (ii) non-behavioral segments show stable movement with dense point distributions, which are less informative.
Inspired by the observation, we first identify behavior segments $\mathcal{A}$ in a flight trajectory $T$ and then adaptively patch the trajectory based on $\mathcal{A}$.

Specifically, considering a flight trajectory indicated by \( T = \{ x_1, x_2, \dots, x_n \} \), we first apply a threshold filter to remove noisy points with excessive oscillations.
Next, we calculate the angle change for each point as: $\text{angle}_i = \{a_1, a_2, \ldots, a_n\}$.

Then, points with angle changes exceeding a threshold $s$ are identified as active points, denoted by $\mathcal{A}'$:

\begin{equation*}
\mathcal{A}' = \{a_j \mid \text{angle}_j > s\}
\end{equation*}

After that, we use a behavior-based patching algorithm to create patches as the inputs of Transformer. 
Since each behavior is composed of multiple consecutive points, for each active point, we compute its index distance to neighboring active points and cluster those with index distance less than a threshold. Then, we select a central active point $c$ from each cluster as a patch center and we can get a patch with a predefined patch size $S$. As illustrated in Figure \ref{fig:pacthing}, $a_{j}$, $a_{j+1}$, and $a_{j+2}$ are all active points. Since they are adjacent, we consider them to be in the same cluster and select point $a_{j+1}$ as the center point $c$. After that, we obtain a behavioral patch set $\mathbf{P}_b$ consists of $g$ patches, where $g$ is the number of active point clusters. Each patch in $\mathbf{P}_b$ preserves the information of a behavioral segment, improving the model’s ability to learn from local patterns.

For non-behavioral segments, we perform uniform sampling with a step size of $\frac{n - S \cdot g}{N-g}$ to generate additional patches from these segments, where the $N$ denotes the number of patches and $S$ is the patch size. Overall, we get a sequence of patches $\mathbf{P}=[p_1, \dots, p_N]$ which consists of $g$ behavioral patches and $N-g$ non-behavioral patches.

\subsubsection{Patch Transformer}

\our utilizes a similar approach similar to PatchTST~\cite{nie2022time} to learn the trajectory embeddings. 


Specifically, we transform each patch $p_i$ into the latent space of dimension $d_m$ via a trainable linear projection $\mathcal{W}_p$, and add a learnable additive position encoding matrix $\mathcal{W}_{pos}$ to encode its temporal order:
$$\mathbf{Y} = \mathbf{W}_p \mathbf{P} + \mathbf{W}_{pos}.$$ 
Then, we feed $\mathbf{Y}$ into a Transformer to generate the final trajectory embedding $\mathbf{v} = \textbf{Transformer}(\mathbf{Y})$, where $\mathbf{v} \in \mathbb{R}^{N \times S \times d_m}$.

\subsection{Optimization of \our}

\begin{figure}[t]
\label{NLL}
\centering
\includegraphics[width=1\columnwidth]{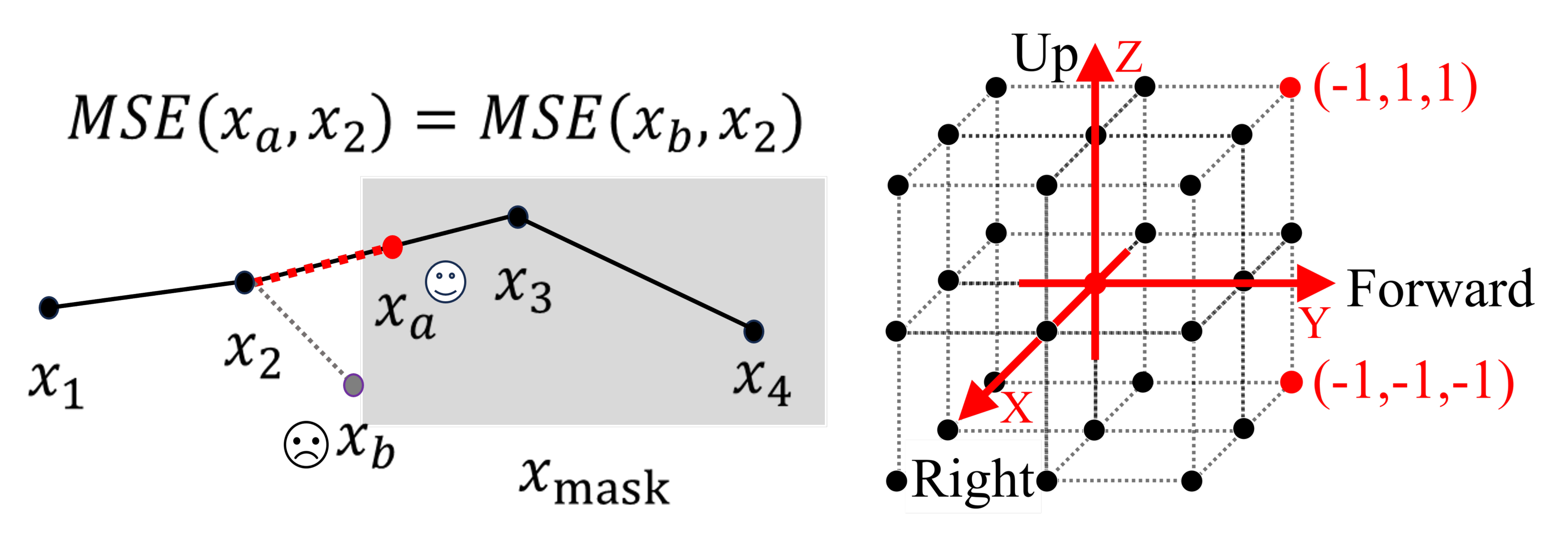} 
\vspace{-3ex}
\caption{Illustration of Motion Trend Learning}
\vspace{-3ex}
\label{fig1}
\end{figure}

To model the complex moving patterns and the spatial continuity in flight trajectories, we optimize \our with a mask-based self-supervised learning strategy and a novel motion trend learning. 

\subsubsection{Mask-based self-supervised learning.} Previous methods often adopt random masking to optimize patch Transformer, which is not suitable for flight trajectories where most points are moving uniformly in a straight line. For these points, the model can easily infer the masked values through simple interpolation between adjacent time points, without capturing the complex trajectory patterns. To address this problem, we introduce a motion-based randomization strategy to effectively mask the behavior patches $\mathbf{P}_b$ and their surrounding areas. Specifically, we mask the patches in $\mathbf{P}_b$ and their neighboring patches with the probability $\rho_b$, and we mask other patches with the probability $\rho_n$, where $\rho_n < \rho_b$. Then, we use a $D \times P$ linear layer to reconstruct the masked patches $x^{mask}$. The Mean Squared Error (MSE) loss to minimize reconstruction errors:
\[
\mathcal{L}_{MSE} = \mathbb{E}_x \frac{1}{M} \sum_{i=1}^{M} \left\| \hat{x}^{masked}_{i} - x^{mask}_{i} \right\|_2^2
\]
where $M$ is the number of points in the masked patches.

\subsubsection{Motion Trend Learning.} However, MSE loss is not enough to model flight trajectories for two main reasons. First, MSE loss treats all errors equally, regardless of their spatial context, which means it does not effectively capture the spatial continuity and dependencies inherent in flight trajectories. Second, MSE loss does not adequately model the uncertainty and sparsity of significant behavioral points within dense trajectory data. Critical behavioral segments, such as turns and altitude changes, are sparse but vital for accurate representation. Optimizing the model with only MSE loss may result in suboptimal learning, as it might not adequately focus on these sparse yet crucial points.

For example, as illustrated in Figure \ref{fig1}, consider a point $x_1$ that needs to be reconstructed or predicted, with two candidate points $x_a$ and $x_b$. Both $x_a$ and $x_b$ are equidistant from $x_1$, resulting in the same MSE loss. But $x_a$ is better than $x_b$ because it is located on the line where the trajectory is moving forward. To model this preference and inherent uncertainty, we introduce the moving direction loss function.

Specifically, we categorize each point in the trajectory based on its movement direction. For point $x_i = (lon_i, lat_i, alt_i)$ and $x_{i+1} = (lon_{i+1}, lat_{i+1}, alt_{i+1})$, we calculate the direction vector $\vec{d}$ from $x_i$ to $x_{i+1}$:
\[
\vec{d} = (lon_{i+1} - lon_i, lat_{i+1} - lat_i, alt_{i+1} - alt_i).
\]
Each component of $\vec{d}$ can be positive, negative, or zero, corresponding to the direction along each axis. The direction is represented as a triplet $(d_{lon}, d_{lat}, d_{alt})$ based on the direction vector, where $d_{lon} = \text{sign}(lon_{i+1} - lon_i)$, $d_{lat} = \text{sign}(lat_{i+1} - lat_i)$, and $d_{alt} = \text{sign}(alt_{i+1} - alt_i)$. The sign function is defined as:
\[
\text{sign}(u) = 
\begin{cases} 
1 & \text{if } u > 0 \\
0 & \text{if } u = 0 \\
-1 & \text{if } u < 0 
\end{cases}
\]
We divided the 3D moving space into 26 categories based on directional movement. Each dimension (longitude, latitude, and altitude) has three possible states: positive, negative, or unchanged. Combining these states yields $3 \times 3 \times 3 - 1 = 26$ unique directions, excluding the case where all dimensions remain unchanged. Each point is then assigned a category corresponding to its movement direction.

We adopt following moving direction loss to optimize the model and to learn these directional preferences:
\begin{equation*}
\mathcal{L}_{MD} = -\mathbb{E}_x \log \mathbb{P}(y_i^{masked} \mid y_{1:i-1}^{mask}, x)
\end{equation*}
where $y_i^{masked}$ represents the label of one of the 26 moving directions.

The final loss is a combination of Mean Squared Error (MSE) loss and the moving direction loss, enabling the model to capture both spatial relationships and precise values. Thus, the final combined loss function is:
\begin{equation*}
\mathcal{L}_{m} = \mathcal{L}_{MD} + \lambda \cdot \mathcal{L}_{MSE}
\end{equation*}
where $\lambda$ is a weighting factor that balances the contributions of the two parts.

By incorporating the spatial proximity-aware loss function and the moving direction loss, \our can effectively encode the complex three-dimensional movement patterns in flight trajectories into the learned representations.

\subsection{Complexity Analysis}

The complexity of the behavior-based patching is $O(n)$, where $n$ represents the length of the flight trajectory. Then, the complexity of patch projection and position embedding is $O(N \cdot S \cdot d_m)$, where $N,S,d_m$ denote the number of patches, the size of each patch and the embedding dimension, respectively. Then, the backbone, \ie, the Transformer encoder, involves self-attention computation, resulting in a complexity of $O(N^2 \cdot d_m)$. Finally, the complexity of the Linear Layer is $O(N \cdot d_m)$. In general, the time complexity of our framework is $O(N^2 \cdot d_m)$, with the Transformer Encoder being the most computationally intensive component. In practice, the number of patches is much less the length of trajectory which makes \our efficient.

\begin{table*}[h!]
\centering
\caption{The experimental results of FTP task.}
\vspace{-2ex}
\small 
\label{table:results}
\renewcommand{\arraystretch}{1.2}
\begin{tabular}{c|c|ccc|ccc|ccc|c}
\hline\hline
\multirow{2}{*}{\textbf{Methods}} & \multirow{2}{*}{\textbf{Hor.}} & \multicolumn{3}{c|}{\textbf{MAE $\downarrow$}} & \multicolumn{3}{c|}{\textbf{MAPE (\%) $\downarrow$}} & \multicolumn{3}{c|}{\textbf{RMSE $\downarrow$}} & \multirow{2}{*}{\textbf{MDE $\downarrow$}} \\ 
 &  & \textbf{Lon} & \textbf{Lat} & \textbf{Alt} & \textbf{Lon} & \textbf{Lat} & \textbf{Alt} & \textbf{Lon} & \textbf{Lat} & \textbf{Alt} &  \\ \hline
  & 1 & 0.0078 & 0.0075 & 2.45 & 0.0062 & 0.0289 & 0.38 & 0.0133 & 0.0148 & 7.18 & 1.28 \\
\multirow{3}{*}{LSTM+Attention} & 3 & 0.0073 & 0.0069 & 3.09 & 0.0058 & 0.0241 & 0.47 & 0.0123 & 0.0146 & 9.86 & 1.23 \\
 & 15 & 0.0142 & 0.0152 & 9.13 & 0.0124 & 0.0543 & 1.97 & 0.0319 & 0.0367 & 20.48 & 1.75 \\
 & 30 & 0.0257 & 0.0296 & 18.66 & 0.0179 & 0.0922 & 3.79 & 0.0592 & 0.0695 & 40.16 & 2.94 \\ 
 & 60 & 0.0735 & 0.0698 & 50.25 & 0.0699 & 0.2314 & 13.78 & 0.1763 & 0.1998 & 102.06 & 6.79 \\ \cline{1-12} 
  & 1 & 0.0041 & 0.0039 & 1.47 & 0.0032 & 0.0136 & 0.32 & 0.0125 & 0.0134 & 8.04 & 0.79 \\
\multirow{3}{*}{PatchTST} & 3 & 0.0073 & 0.0072 & 2.98 & 0.0063 & 0.0256 & 0.61 & 0.0214 & 0.0231 & 8.72 & 1.34 \\
 & 15 & 0.0126 & 0.0129 & 8.03 & 0.0125 & 0.0532 & 1.77 & 0.0305 & 0.0322 & 19.94 & 1.61 \\
 & 30 & 0.0231 & 0.0248 & 14.99 & 0.0217 & 0.0962 & 3.48 & 0.0591 & 0.0672 & 38.88 & 2.70 \\ 
 & 60 & 0.0555 & 0.0572 & 24.65 & 0.0489 & 0.1978 & 7.87 & 0.1129 & 0.1376 & 72.15 & 6.50 \\ \cline{1-12}
  & 1 & 0.0018 & 0.0018 & 1.16 & 0.0016 & 0.0067 & 0.21 & 0.0037 & 0.0116 & 13.11 & 0.32 \\
\multirow{3}{*}{FlightBERT++} & 3 & 0.0032 & 0.0032 & 2.33 & 0.0031 & 0.0119 & 0.43 & 0.0076 & 0.0133 & 12.88 & 0.59 \\
 & 15 & 0.0125 & 0.0118 & 7.48 & 0.0110 & 0.0429 & 1.38 & 0.0268 & 0.0329 & 22.91 & 1.99 \\
 & 30 & 0.0214 & 0.0213 & 12.97 & 0.0286 & 0.0916 & 2.46 & 0.0576 & 0.0671 & 48.55 & 4.44 \\ 
 & 60 & 0.0482 & 0.0497 & 29.87 & 0.0606 & 0.1666 & 6.02 & 0.1076 & 0.1319 & 79.25 & 7.24 \\ \cline{1-12} 
  & 1 & 0.0019 & 0.0018 & 1.17 & 0.0015 & 0.0062 & 0.23 & 0.0038 & 0.0121 & 8.02 & 0.34 \\
\multirow{3}{*}{\our} & 3 & 0.0034 & 0.0033 & 2.32 & 0.0029 & 0.0118 & 0.42 & 0.0075 & 0.0136 & 8.11 & 0.57 \\
 & 15 & 0.0124 & 0.0121 & 7.58 & 0.0109 & 0.0431 & 1.36 & 0.0259 & 0.0319 & \textbf{18.24} & 1.98 \\
 & 30 & 0.0196 & 0.0213 & 12.14 & 0.0212 & 0.0811 & 2.09 & 0.0436 & 0.0501 & 37.52 & 3.34 \\ 
 & 60 & 0.0381 & 0.0401 & 25.67 & 0.0403 & 0.1216 & 3.72 & 0.0836 & 0.1019 & 56.45 & 5.14 \\
 \cline{1-12}
  & 1 & \textbf{0.0017} & \textbf{0.0017} & \textbf{1.15} & \textbf{0.0015} & \textbf{0.0061} & \textbf{0.19} & \textbf{0.0036} & \textbf{0.0112} & \textbf{8.01} & \textbf{0.29} \\
\multirow{3}{*}{\our+BE} & 3 & \textbf{0.0029} & \textbf{0.0029} & \textbf{1.92} & \textbf{0.0028} & \textbf{0.0116} & \textbf{0.39} & \textbf{0.0036} & \textbf{0.0109} & \textbf{7.83} & \textbf{0.55} \\
 & 15 & \textbf{0.0122} & \textbf{0.0112} & \textbf{7.43} & \textbf{0.0101} & \textbf{0.0402} & \textbf{1.31} & \textbf{0.0243} & \textbf{0.0312} & 19.81 & \textbf{1.44} \\
 & 30 & \textbf{0.0192} & \textbf{0.0198} & \textbf{11.89} & \textbf{0.0203} & \textbf{0.0699} & \textbf{2.03} & \textbf{0.0432} & \textbf{0.0498} & \textbf{33.88} & \textbf{2.24} \\ 
 & 60 & \textbf{0.0371} & \textbf{0.0392} & \textbf{25.45} & \textbf{0.0401} & \textbf{0.1209} & \textbf{3.62} & \textbf{0.0819} & \textbf{0.0932} & \textbf{54.77} & \textbf{5.04} \\ \hline \hline

\end{tabular}
\end{table*}

\section{Experiment}

\subsection{Experimental Settings}
In this section, we briefly introduce the datasets, experiment protocols, baselines and hyperparameter settings. The code and data are public available at https://github.com/liushuoer/FLIGHT2VEC. More details of the experimental settings are described in the Appendix.
\subsubsection{Data Descriptions} 
We conduct extensive experiments on two real-world datasets, the Swedish Civil Air Traffic Control (SCAT)~\cite{nilsson2023swedish} and Aircraft Trajectory Classification Data for Air Traffic Management(ATFMTraj)~\cite{ATFMTraj2024}. 

\subsubsection{Experimental Protocol} We employ three representative tasks, \ie flight trajectory prediction (FTP), flight recognition (FR), and anomaly detection (AD), to evaluate the representations learned from \our. For FTP, we predict the future trajectory in (1, 3, 15, 30, 60) different horizons and utilize Mean Absolute Error (MAE), Mean
Absolute Percentage Error (MAPE), Root Mean Squared Error (RMSE) and Mean Distance Error (MDE) as the evaluation metrics. For FR, we directly use the category of flight as ground truth and evaluate the performance with accuracy (ACC), precision (PRE), and recall (REC). For AD, we generate the synthetic anomalies according to the previous work~\cite{guo2022data} and evaluate the performance of \our with AUC and AUPR. Following the settings of the previous work, we use the Mean Time Cost(MTC) metric to evaluate the computational performance.

\begin{table*}[h!]
    \centering
    \caption{The experimental results of anomaly detection task.}\label{tab:ad}
    \vspace{-2ex}
    \begin{tabular}{c|cccc|cccc}
    \hline\hline
     & \multicolumn{4}{c|}{\textbf{AUC}} & \multicolumn{4}{c}{\textbf{AUPR}} \\
    \multirow{-2}{*}{Method} & SMA & HD & VD & Go-Around & SMA & HD & VD & Go-Around \\ \hline
    \multicolumn{1}{c|}{DMDN} & 0.7498 & 0.7444 & 0.7354 & 0.7100 & 0.7416 & 0.7498 & 0.7326 & 0.7210 \\
    \multicolumn{1}{c|}{DDM} & 0.8509 & 0.8411 & 0.8324 & 0.8200 & 0.8122 & 0.8112 & 0.7923 & 0.7890 \\
    \multicolumn{1}{c|}{\our} & \textbf{0.9201} & \textbf{0.9178} & \textbf{0.9127} & \textbf{0.9050} & \textbf{0.9072} & \textbf{0.9059} & \textbf{0.9017} & \textbf{0.9005} \\ \hline
    \hline
    \end{tabular}
\end{table*}

\begin{table*}[h!]
\centering
\caption{The experimental results of flight recognition task.}
\vspace{-2ex}
\small 
\renewcommand{\arraystretch}{1.2}
\label{table:recognition}
\begin{tabular}{l|ccc|ccc|ccc|ccc}
\hline
\hline
\multirow{2}{*}{\textbf{Methods}} & \multicolumn{3}{c|}{\textbf{RKSla}} & \multicolumn{3}{c|}{\textbf{RKSId}} & \multicolumn{3}{c|}{\textbf{ESSA}} & \multicolumn{3}{c}{\textbf{LSZH}} \\ 
 & \textbf{ACC} & \textbf{PRE} & \textbf{REC} & \textbf{ACC} & \textbf{PRE} & \textbf{REC} & \textbf{ACC} & \textbf{PRE} & \textbf{REC} & \textbf{ACC} & \textbf{PRE} & \textbf{REC} \\ \hline
 SPIRAL & 0.8344 & 0.8301 & 0.8302 & 0.9946 & 0.9942 & 0.9932 & 0.8503 & 0.8496 & 0.8408 & 0.9173 & 0.9172 & 0.9124 \\ 
 ATSCC & 0.9946 & 0.9951 & 0.9911 & 0.9987 & \textbf{0.9981} & 0.9971 & 0.9990 & \textbf{0.9989} & 0.9980 & 0.9977 & 0.9979 & 0.9971 \\
 Flight2Vec & \textbf{0.9947} & \textbf{0.9952} & \textbf{0.9943} & \textbf{0.9988} & 0.9980 & \textbf{0.9970} & \textbf{0.9990} & 0.9986 & \textbf{0.9982} & \textbf{0.9981} & \textbf{0.9980} & \textbf{0.9980} \\ \hline
\hline
\end{tabular}
\vspace{-3ex}
\end{table*}

\subsubsection{Baselines} 
As described in the experimental protocol, we employ three tasks to verify the performance of \our. For flight trajectory prediction, three representative methods, \ie \textbf{FlightBERT++}~\cite{guo2024flightbert++},  \textbf{LSTM+Attention}~\cite{guo2022flightbert}, and \textbf{PatchTST}~\cite{nie2022time} are compared. For anomaly detection methods, we use \textbf{DMDN}~ \cite{lijing2021aircraft}, and \textbf{DDM}~\cite{guo2022data} for performance comparison. For flight trajectory recognition, we select \textbf{SPIRAL}~\cite{lei2019similarity} and \textbf{ATSCC}~\cite{ATSCC2024} as our baselines. 
Moreover, we also compare the ablations of \our to verify the superiority of the proposed techniques and study the parameter sensitivity to provide some insight in the use of \our.

\subsubsection{Hyperparameters setting} 
Our model utilizes the Transformer configuration from~\cite{nie2022time}, which includes $3$ layers with a model dimension of $256$, and $16$ attention heads with a dropout rate of $0.2$. The binomial masking probability is set at $0.4$. The dimension of the representation $p_{i}$ is set to $256$. For training, the batch size is set to $256$, and the AdamW optimizer is used with a learning rate of $1 \times 10^{-5}$ The model is pre-trained for $100$ epochs with a patch length of $32$. All the experiments are conducted on the $2\times$ NVIDIA 3090Ti. 

\subsection{Effectiveness of \our}
To evaluate the effectiveness of \our, we conduct experiments on the aforementioned three tasks and analyze the results of each task respectively.

\subsubsection{Results of Flight Trajectory Prediction.} We conduct FTP experiment on SCAT dataset and report the quantitative results in Table \ref{table:results}. According to the results, we have four observations. Firstly, the Transformer-based methods, such as PatchTST, FlightBERT++ and \our, achieve superior performance compared with the LSTM+Attention. Owing to the large parameter size and self-attention mechanism, Transformer-based methods have higher model capacity to model the long-term dependence for flight prediction. Secondly, FlightBERT++ is the most competitive baseline, which can predict the flight trajectory in a non-autoregressive manner, but it is also inferior to \our. This proves the effectiveness of activity density and moving trends modeled in \our. Thirdly, with the prediction horizon increasing, the performances of all methods are dropped. For example, the MDE of FlightBERT++ increases by approximately $237$\% when the horizon increases from $3$ to $15$. \our still beats all compared baselines on all prediction horizons. We attribute this to the effectiveness of moving direction loss which captures the 3D spatial motion trend of flight trajectory. Lastly, due to the fusion of some attributes of the trajectory points, FlightBERT++ achieves better performance than \our for short-term horizon predictions (in horizons $1$ and $3$). We use the same method in FlightBERT++ to integrate the attributes in \our and form \our+BE. As shown, \our+BE achieves the best performance across all prediction horizons, which also proves the scalability of our method. 

\subsubsection{Results of Anomaly Detection.} The flight trajectory anomaly detection experiments are also conducted on SCAT dataset. The results are presented in Table \ref{tab:ad}. Due to the absence of real ground truth anomalies. We construct four types of anomalies, \ie, Successive Multipoint Anomaly (SMA), Horizontal Deviation (HD), Vertical Deviation (VD) and Go-Around following the existing work~\cite{guo2022data}. We train another MLP layer along with the learned representations of \our to achieve anomaly detection. All the methods are trained with the same training dataset. For evaluation metrics, we utilize  AUPR in addition to AUC to handle the unbalanced anomaly classes. As shown, the performance of DMDN is inferior to DDM, which proves the density estimation method of DMDN can not learn representative features for detecting anomalies. \our achieves significant improvements in all metrics, proving that the learned representations can capture the local movement patterns of the flight trajectory. For Go-Around anomaly, \our achieves  $10.4$\% and $14.1$\% improvements of AUC and AUPR, compared with the most competitive baseline DDM. Moreover, the results of \our consistently outperform the baselines on four anomaly types,  indicating that the proposed behavior-adaptive patching and moving direction loss can encourage the representation model to learn useful features.

\subsubsection{Results of Flight Recognition.} We conduct the flight recognition experiment on ATFMTraj dataset and present the results in Table \ref{table:recognition}. In ATFMTraj, the flight trajectories are categorized into four classes, \ie, RKSla, RKSId, ESSA and LSZH, by the airports of take-off and landing. As illustrated, the performances of all methods are relatively high. The best accuracy of the four classes are over $99$\%. For the compared baselines, the performance of ATSCC significantly outperforms SPIRAL. The precision increased from $0.8301$ to $0.9946$ on RKSla. This phenomenon indicates the segment-patch framework in ATSCC can better represent the flight trajectory compared with using original data directly. \our utilizes the behavior-adaptive patching Transformer and achieves comparable results with the performance of ATSCC.

\subsection{Efficiency of \our}
To verify the efficiency of \our, we report the Mean Time Cost (MTC) of representation generation along with the model size in Table ~\ref{table:efficiency}. As shown, the computational time of \our is better than, at least comparable with, all the compared baselines. For the FTP task, \our is the fast method expect for PatchTST. However, PatchTST is a light model that is not specifically designed for flight trajectories and performs poorer than \our. FlightBERT++ is inferior to \our, because FlightBERT++ is a encoder-decoder framework while \our is a decoder-only architecture. The parameters in \our are much less than FlightBERT++, leading to better computational efficiency. For FR, SPIRAL has minimal computational time. This is because SPIRAL is not a deep learning method. Compared with the SOTA method, \our is 10 times faster than ATSCC, \ie, from $43.89$ ms (ATSCC) to $3.11$ ms (\our).  For AD task, \our demonstrate substantial improvements in computational performance compared to the baselines.

\begin{table}[!tb]
\centering
\caption{Comparison of the Computational Performance.}
\vspace{-2ex}
\small 
\label{table:efficiency}
\renewcommand{\arraystretch}{1.2}
\begin{tabular}{c|c|c|c}
\hline\hline
\textbf{Task} & \textbf{Methods} & \textbf{Parameters (M)} & \textbf{MTC (ms)} \\ \hline

\multirow{4}{*}{FTP} 
 & LSTM+Attention & 0.90 & 48.96 \\
 & PatchTST & 1.61 & 2.20 \\
 & FlightBERT++ & 29.96 & 7.21 \\
 & \our & 1.86 & 3.12 \\ \hline
 
\multirow{3}{*}{FR} 
 & SPIRAL & -- & 0.76 \\
 & ATSCC & 0.90 & 43.89 \\
 & \our & 1.85 & 3.11 \\ \hline

\multirow{3}{*}{AD} 
 & DMDN & 0.61 & 47.96 \\
 & DDM & 0.11 & 6.53 \\
 & \our & 1.85 & 3.11 \\ \hline\hline

\end{tabular}

\end{table}

\begin{table}[!tb]
\centering
\caption{The experimental results of the ablation study.}
\vspace{-2ex}
\scriptsize 
\label{table:ablate}
\renewcommand{\arraystretch}{1.2}
\setlength{\tabcolsep}{2pt} 
\begin{tabular}{c|c|ccc|ccc|ccc}
\hline\hline
\multirow{2}{*}{\textbf{Methods}} & \multirow{2}{*}{\textbf{Hor.}} & \multicolumn{3}{c|}{\textbf{MAE $\downarrow$}} & \multicolumn{3}{c|}{\textbf{MAPE (\%) $\downarrow$}} & \multicolumn{3}{c}{\textbf{RMSE $\downarrow$}} \\ 
 &  & \textbf{Lon} & \textbf{Lat} & \textbf{Alt} & \textbf{Lon} & \textbf{Lat} & \textbf{Alt} & \textbf{Lon} & \textbf{Lat} & \textbf{Alt} \\ \hline
\multirow{3}{*}{w/o PD} 
 & 1 & 0.0032 & 0.0032 & 1.65 & 0.0029 & 0.0098 & 0.54 & 0.0052 & 0.0145 & 14.55 \\
 & 15 & 0.0119 & 0.0117 & 7.94 & 0.0117 & 0.0512 & 1.62 & 0.0289 & 0.0276 & 19.12 \\ \hline
\multirow{3}{*}{w/o MD} 
 & 1 & 0.0022 & 0.0022 & 1.39 & 0.0019 & 0.0072 & 0.29 & 0.0042 & 0.0133 & 13.95 \\
 & 15 & 0.0144 & 0.0141 & 7.88 & 0.0162 & 0.0438 & 1.42 & 0.0279 & 0.0382 & 23.11 \\ \hline
\multirow{3}{*}{our} 
 & 1 & 0.0019 & 0.0018 & 1.17 & 0.0015 & 0.0062 & 0.23 & 0.0038 & 0.0121 & 13.21 \\
 & 15 & 0.0124 & 0.0121 & 7.58 & 0.0109 & 0.0431 & 1.36 & 0.0259 & 0.0319 & 18.24 \\ \hline \hline
\end{tabular}
\vspace{-4ex}
\end{table}

\subsection{Ablation Study}
We compare \our with two ablations to analyze the effectiveness of the proposed components. We remove the proposed behavior-based patching, randomly sample points at fixed intervals and divide the patch to obtain w/o PD. We obtain w/o MD by removing moving direction loss.

Due to the space limit, we only report the experiment on FTP task and the results are shown in Table \ref{table:ablate}. We observe: (1) Comparing the results of \our with w/o MD, we observe the moving direction loss can model the spatial continuity flight trajectories. For example, the RMSE improves from $18.24$ to $23.11$ on altitude. (2) From the results of w/o PD and \our, we can conclude that the activity patching mechanism capacity to represent flight trajectory is more effective than a fixed patch. (3) \our achieves the best performance compared to all ablations, which proves the effectiveness of the proposed techniques.

\subsection{Sensitivity Analysis}

\begin{figure}[t]
\centering
\includegraphics[width=0.9\columnwidth]{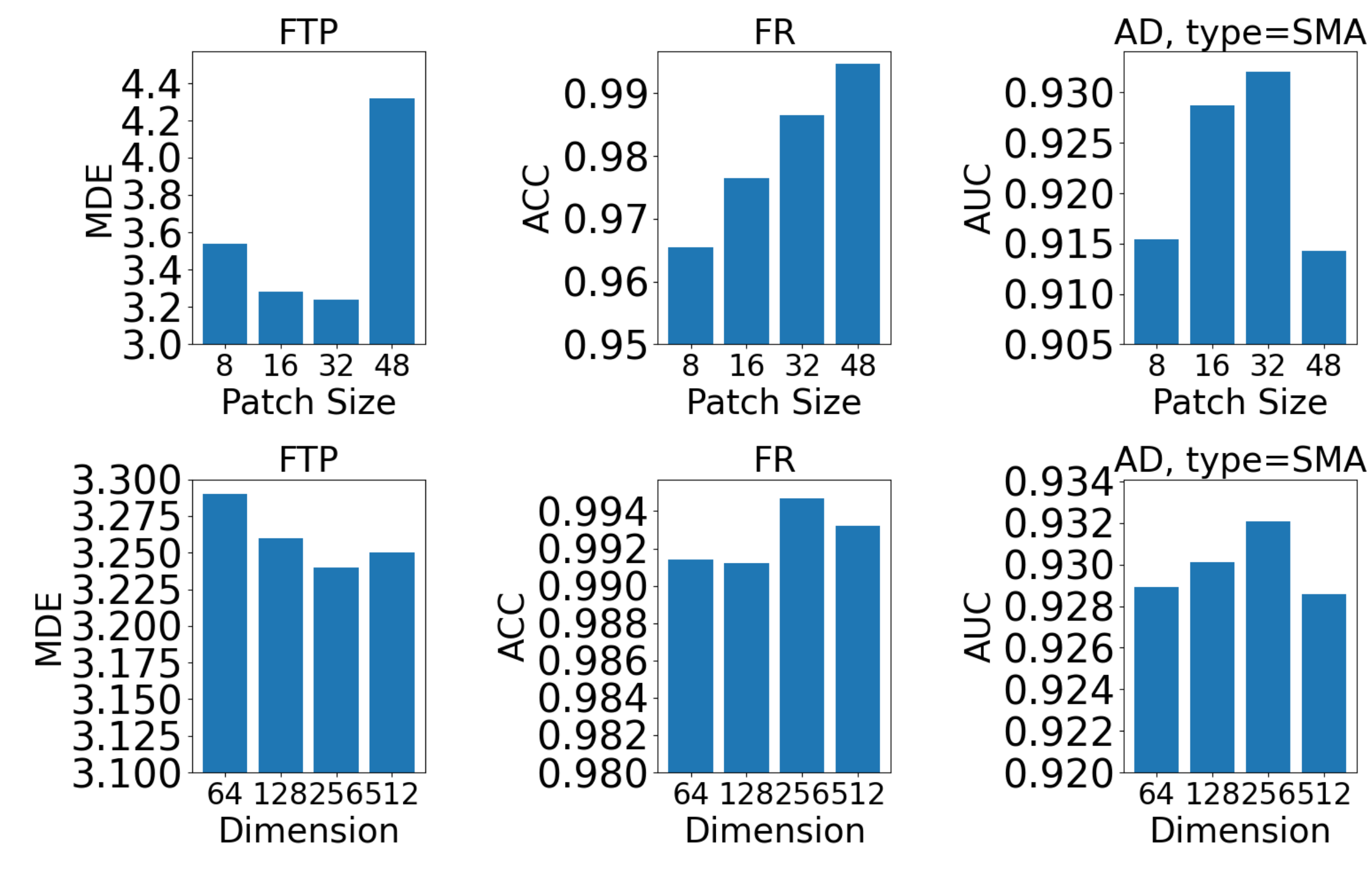} 
\vspace{-2ex}
\caption{MDE scores with varying model parameters.}
\vspace{-4ex}
\label{fig:sen}
\end{figure}

We analyze the impact of selecting different patch sizes and representation dimensions on different tasks. We ranged the patch size from 8 to 48 and the representation dimension from 64 to 512, and then show the average performance of these configurations on different tasks in Figure \ref{fig:sen}. 

With the increase of patch size, the performance of \our first increases and then drops. \our achieves the best performance with the patch size of $32$ on both FTP and AD tasks.  For FR task, the optimal performance is achieved with the largest patch size of $48$. This is because the FR task pays more attention to the global movement patterns of the flight trajectory.

For the change of embedding dimension, the performances of the three tasks have similar patterns. \our achieves the best performance on the dimension size of $256$. This phenomenon indicates \our is robust to different embedding dimension sizes.

\subsection{Conclusion}
In this paper, we present \our which is the first unified framework specifically designed for flight trajectory representation learning. By addressing the challenges of unbalanced behavior density and 3D spatial continuity, \our enhances the representation of flight trajectories. Our approach introduces an adaptive behavior-based patching mechanism and a direction loss, significantly improving performance across tasks like trajectory prediction, anomaly detection, and flight monitoring. Extensive experiments across multiple task demonstrate that \our not only significantly outperforms existing methods but also sets a new benchmark in the field.

\newpage
\section*{Acknowledgment}
This work has been supported by the National Natural Science Foundation of China under Grant No. 62472405. This work is also sponsored by CCF-DiDi GAIA Collaborative Research Funds for Young Scholars.
\bibliography{aaai25}
\newpage
\section*{Appendix}
\subsection{Details of datasets.}
\begin{itemize}
    \item \textbf{SCAT:} The SCAT dataset contains detailed data of almost 170,000 flights from October 2016 to September 2017, as well as weather forecasts and airspace data collected from the air traffic control system in the Swedish flight information region. The dataset is limited to scheduled flights, excluding military and private aircraft. After removing trajectories that only flew at a single altitude, we obtained a final dataset of $100,383$ trajectories.
    \item \textbf{ATFMTraj:} The dataset has classification labels based on aeronautical publications. The dataset is divided into four sub-datasets based on the airport and flight plan (departing or arriving). The arrival and departure datasets of Incheon International Airport (ICAO code: RKSI) are denoted as RKSIa and RKSId, respectively. The arrival datasets of Stockholm Arlanda Airport (ICAO code: ESSA) and Zurich Airport (ICAO code: LSZH) are denoted as ESSA and LSZH, respectively.
\end{itemize}

\subsection{Details of Compared Baselines.}
We compare \our with seven baselines, which are divided into three groups based on downstream tasks: flight trajectory prediction (FTP) methods, flight trajectory recognition methods, and anomaly detection methods.  For the first group, we employ three baselines. We compare the \textit{State-of-the-Art} (SOTA) model \textbf{FlightBert++}\cite{guo2024flightbert++} in the field of flight trajectory prediction and the model \textbf{LSTM+Attention}\cite{guo2022flightbert} based on the Seq2Seq architecture according to its experimental settings. We also compare the patch-based Transformer model  \textbf{PatchTST}\cite{nie2022time} and the model \textbf{\our+BE} that introduces the additional features proposed by \textbf{FlightBert++}. For flight trajectory recognition methods, we select \textbf{SPIRAL}\cite{lei2019similarity} and \textbf{ATSCC}\cite{ATSCC2024}, as our baselines. For anomaly detection methods, we use \textbf{DMDN} \cite{lijing2021aircraft}, and \textbf{DDM} \cite{guo2022data} for performance comparison.

\subsection{Experimental Protocol.} In our experiments, each dataset is divided into two subsets: the first $50\%$ of timestamps is denoted as the training set, while the latter $50\%$  is denoted as the test set. We utilize three downstream tasks to evaluate the performance of \our. For trajectory prediction task, we predicts the results in a single inference process for multiple time steps on the SCAT data set. The experimental results are divided into five (1, 3, 15, 30, 60) different horizons, corresponding to 20 seconds, 1, 5, 10 and 20 minutes trajectories in the future. For flight trajectory recognition task, we directly used the real label to conduct the experiments. Due to privacy concerns, Existing flight trajectory dataset either have no labeled anomaly edges or only have one anomaly type. To verify the ability of \our on various anomaly types, we follow the experiments of \cite{guo2022data} and generate three kinds of systematic anomaly types, \ie, Successive Multipoint Anomaly (SMA), Horizontal Deviation (HD), Vertical Deviation (VD) and Go-Around for SCAT datasets. SMA refers to multiple consecutive trajectory points in a flight that deviate from the planned flight path. This anomaly may indicate a problem with the aircraft's control or navigation. HD refers to the aircraft's deviation from the planned horizontal path during flight. Especially during the approach phase, HD anomalies may affect flight safety. VD refers to the aircraft's deviation from the expected vertical flight trajectory. The aircraft fails to accurately follow the glide path, which may cause the aircraft to be in a dangerous state of being too high or too low during approach or landing. Go-Around refers to aborting the landing and re-entering the route when the aircraft cannot land safely. In the anomaly detection task, in order to simulate the situation with few anomaly in real-world scenarios, we only add $5\%$ anomaly lable data to the test set.

\end{document}